\documentclass[10pt,twocolumn,letterpaper]{article}

\usepackage{cvpr}      

\usepackage{graphicx}
\usepackage{amsmath}
\usepackage{amssymb}
\usepackage{booktabs}
\usepackage{multirow}
\usepackage{soul}

\newcommand{\display}[1]{{\iffalse{#1}\fi}} 

\newcommand{\jc}{\textcolor[rgb]{0,0.5,0}}

\usepackage[pagebackref,breaklinks,colorlinks]{hyperref}

\usepackage[capitalize]{cleveref}
\crefname{section}{Sec.}{Secs.}
\Crefname{section}{Section}{Sections}
\Crefname{table}{Table}{Tables}
\crefname{table}{Tab.}{Tabs.}

\def\confName{CVPR}

\begin{document}

\title{Weakly-Supervised Temporal Action Localization\\
by Progressive Complementary Learning}

\author{Jia-Run Du$^{1}$, Jia-Chang Feng$^{1,2}$, Kun-Yu Lin$^{1}$, Fa-Ting Hong$^{1}$, Xiao-Ming Wu$^{1}$ \\ Zhongang Qi$^{2,3}$, Ying Shan$^{2,3}$, Wei-Shi Zheng$^{1}$ \\
$^{1}$ School of Computer Science and Engineering, Sun Yat-sen University \\
$^{2}$ ARC Lab, $^{3}$ Tencent PCG\\
}
\maketitle




\begin{abstract}
   Weakly Supervised Temporal Action Localization (WSTAL) aims to localize and classify action instances in long untrimmed videos with only video-level category labels. Due to the lack of snippet-level supervision for indicating action boundaries, previous methods typically assign pseudo labels for unlabeled snippets. However, since some action instances of different categories are visually similar, it is non-trivial to exactly label the (usually) one action category for a snippet, and incorrect pseudo labels would impair the localization performance. To address this problem, we propose a novel method from a category exclusion perspective, named Progressive Complementary Learning (ProCL), which gradually enhances the snippet-level supervision. Our method is inspired by the fact that video-level labels precisely indicate the categories that all snippets surely do not belong to, which is ignored by previous works. Accordingly, we first exclude these surely non-existent categories by a complementary learning loss. And then, we introduce the background-aware pseudo complementary labeling in order to exclude more categories for snippets of less ambiguity. Furthermore, for the remaining ambiguous snippets, we attempt to reduce the ambiguity by distinguishing foreground actions from the background. Extensive experimental results show that our method achieves new state-of-the-art performance on two popular benchmarks, namely THUMOS14 and ActivityNet1.3.
   
\end{abstract}

\section{Introduction}
\label{sec:intro}
\begin{figure}
    \centering
    \includegraphics[width=\linewidth]{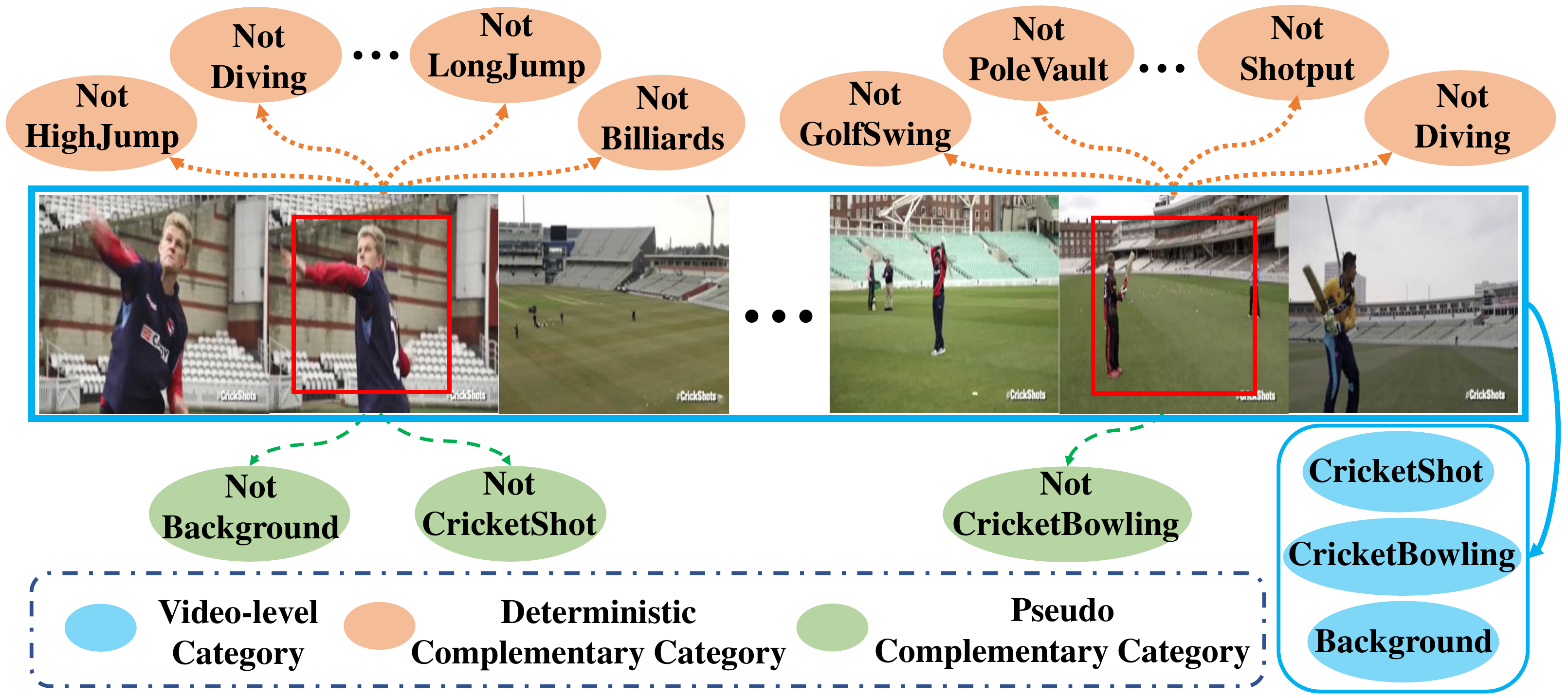}
    \caption{Illustration of \textit{complementary categories}. Given a video containing background and two action categories, \ie, ``CricketShot" and ``CricketBowling", it is clear that all snippets in the video surely do not belong to the categories ``HighJump", ``Diving", \etc, which are termed \textit{deterministic complementary categories}. Furthermore, for some snippets (red boxes), we would confidently exclude some categories that these snippets likely do not belong to, which are termed \textit{pseudo complementary categories} (\eg, in the right red box, we easily exclude the category ``CricketBowling'' according to the 
    visible cricket bat). 
    }
    \label{fig:one}
    \vspace{-6mm}
\end{figure}
Temporal action localization aims to localize and classify action instances in an untrimmed video. It is important for applications of video retrieval~\cite{ciptadi2014movement,ramezani2016review}, anomaly detection~\cite{feng2021mist} and highlight detection~\cite{hong2020mini}. 
Fully-supervised temporal action localization~\cite{shou2016temporal,zhao2017temporal,lin2018bsn,lin2019bmn,zhang2019learning} requires the start and end timestamps of each action instance for training.
It is time and labor-consuming to annotate all action instances in long untrimmed videos. 

In this work, we focus on Weakly-Supervised Temporal Action Localization (WSTAL)~\cite{lee2020background,lee2021weakly,hong2021cross,islam2020weakly,islam2021hybrid,liu2021acsnet,liu2019completeness,liu2019learning,ma2020sf,luo2020weakly,narayan20193c,nguyen2018weakly,yang2021uncertainty,zhang2021cola,zhai2020two,zeng2019breaking,zhong2018step,ji2021weakly}, which does not require precise boundary annotations
but only the action categories in the video. 
Typically, existing methods~\cite{lee2021weakly,zhai2020two,islam2021hybrid,nguyen2018weakly,zhang2021cola,luo2020weakly,lee2020background,hong2021cross} exploit the Multi-Instance Learning (MIL) paradigm~\cite{zhou2004multi}, where an untrimmed video is considered as a labeled bag of snippet samples and video-level prediction is aggregated from snippets.
Despite the progress of MIL-based methods, the absence of snippet-level supervision signal still blocks the weakly-supervised temporal action localization. 
However, without the guidance of snippet-level labels, MIL-based methods are prone to localize the most discriminative ~\cite{islam2021hybrid} of action instances.

To alleviate this problem, various methods~\cite{islam2020weakly,zhang2021cola,gong2020learning,lee2021weakly,pardo2021refineloc,luo2020weakly,huang2022multi,he2022asm,zhai2020two,islam2021hybrid,shi2020weakly,li2022forcing,liu2019completeness,paul2018w,wang2017untrimmednets} are proposed, \eg erasing-based~\cite{islam2021hybrid,zeng2019breaking,zhong2018step}, metric-based~\cite{islam2020weakly,zhang2021cola}, feature-based~\cite{hong2021cross}.
Among these solutions, one of the most effective types is based on pseudo labels~\cite{pardo2021refineloc,luo2020weakly,lee2021weakly,zhai2020two,he2022asm}, which assigns pseudo labels for unlabeled snippets. 
For example, Huang \etal~\cite{huang2022multi,zhai2020two} assign pseudo labels by thresholding temporal attention scores with a pre-set threshold, and He \etal~\cite{pardo2021refineloc,he2022asm} assign pseudo labels based on predicted action proposals. However, some action instances of different categories look visually similar, \eg, snippets of ``ThrowDiscus” and ``Shotput”. Therefore, it is non-trivial to exactly label (usually) one action category for a snippet, and incorrect pseudo labels would impair the localization performance.

To address this problem, we note a fact that, the video-level category labels precisely indicate categories that all snippets surely do not belong to. As shown in \Cref{fig:one}, given a video with video-level labels ``CricketShot" and ``CricketBowling", we know that all snippets in the video surely do not belong to ``Diving", ``HighJump", ``GolfSwing", \etc. Thus, we can label all snippets as ``Not Diving”, ``Not HighJump", ``Not GolfSwing", \etc. Moreover, for each snippet, it is intuitively easier to exclude some non-existent categories than exactly predicting the (usually) one ground-truth category (\ie, excluding all non-existent categories). As the snippet marked by the right red box in \cref{fig:one}, it is difficult to accurately predict its action category, since the visually small athletes lead to confusion between the action and background. From another perspective, it is easier to confirm that this snippet does not belong to ``CricketBowling", because the athletes in this snippet hold the cricket bat. 

Accordingly, in this work, we propose a
novel method, named \textbf{Pro}gressive \textbf{C}omplementary \textbf{L}earning (ProCL). ProCL gradually excludes the categories that snippets should not belong to, which we term \textit{complementary categories}. With Complementary Learning loss, we first exclude the categories that all snippets indicated by the video-level labels surely do not belong to, which we term \textit{deterministic complementary categories.} Then, to further enhance the snippet-level supervision, we attempt to exclude more categories than those deterministic complementary categories. After identifying snippets of less ambiguity, we assign pseudo complementary labels to these snippets, aiming to exclude the categories that snippets do likely not belong to, which we term \textit{pseudo complementary categories}. For the remaining ambiguous snippets, we propose to reduce the ambiguity by a Foreground Background Discrimination loss, which coarsely provides further snippet-level supervision by distinguishing foreground actions from the background.
Furthermore, considering the information collaboration between multi-scale snippet sequences, we propose the Multi-scale Pseudo Complementary Learning loss to improve the quality of pseudo complementary labels.

Extensive experiments on THUMOS14 \cite{THUMOS14} and ActivityNet1.3 \cite{caba2015activitynet} datasets demonstrate the superiority of our ProCL over the state-of-the-art methods. We also conduct ablation studies on each component, which verify the rationality of progressive complementary learning. Moreover, our method significantly outperforms existing pseudo-label-based methods~\cite{he2022asm} in providing snippet-level supervision.

\begin{figure*}[t]
    \centering
    \vspace{-5mm}
    \includegraphics[width=0.98\linewidth]{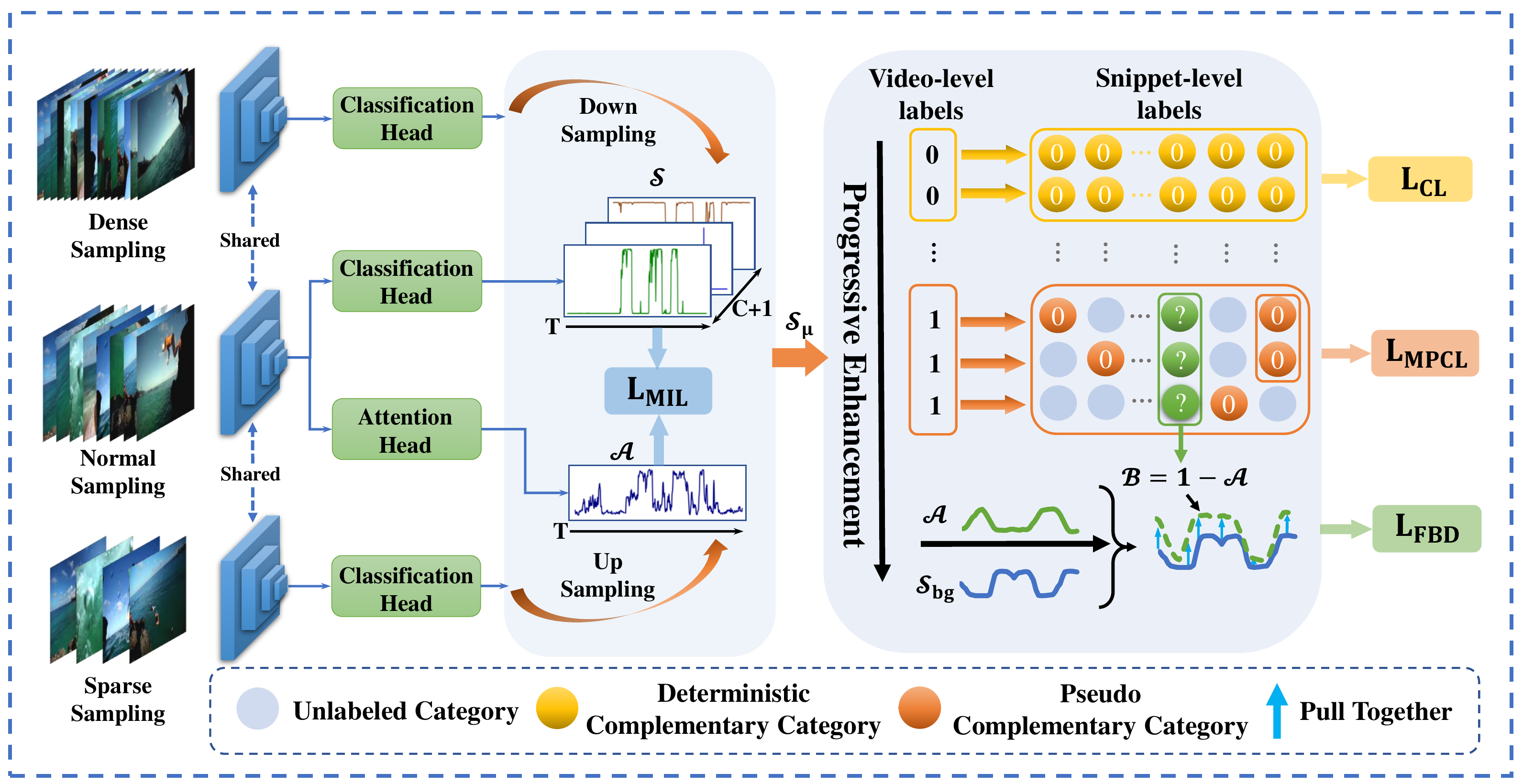}
    \caption{Overview of our proposed Progressive Complementary Learning (ProCL). Given a video, the class activation sequence $\mathcal{S}$ and the class-agnostic attention scores $\mathcal{A}$ are generated by the classification and the attention heads. ProCL progressively excludes categories that snippets should not belong to, for gradually enhancing snippet-level supervision. 
    First, according to the video-level labels, we exclude the categories that all snippets surely do not belong to (\ie, \textit{deterministic complementary categories}) by the Complementary Learning loss $L_{CL}$.
    Then, for snippets of less ambiguity, we exclude some categories that the snippets likely do not belong to (\ie, \textit{pseudo complementary categories}) by the Multi-scale Pseudo Complementary Learning loss $L_{MPCL}$.
    Furthermore, for the remaining ambiguous snippets, we disambiguate them by the Foreground Background Discrimination loss $L_{FBD}$, which aims to coarsely provide further snippet-level supervision.
    Besides, the Multiple-Instance Learning loss $L_{MIL}$ is adopted for video-level supervision.
    }
    \label{fig:two}
    \vspace{-5mm}
\end{figure*}

\section{Related works}
\label{sec:relate}
\noindent\textbf{Fully-Supervised Temporal Action Localization}
aims to find the temporal intervals of action instances from long untrimmed videos and classify them. \jc{To address it}, accurate timestamp annotation for each action instance is required in the given video during training. Several large-scale datasets have been collected for this task, \eg, THUMOS14 \cite{THUMOS14} and ActivityNet \cite{caba2015activitynet}. Many methods \cite{shou2016temporal,zhao2017temporal,lin2018bsn,gao2017turn,lin2019bmn,zeng2019graph,xu2020g} are based on the  proposal-then-classification paradigm, which generate proposals via bottom-up \cite{zhao2017temporal,lin2018bsn,lin2019bmn,zeng2019graph,xu2020g,qing2021temporal} or top-down \cite{shou2016temporal,gao2017turn} manner followed by multi-class classification. Some methods explore a one-stage paradigm, which directly generates class-aware proposals \cite{lin2017single,long2019gaussian,yang2020revisiting}, or group the action and boundary predictions \cite{yuan2017temporal}.

\noindent\textbf{Weakly-Supervised Temporal Action Localization} aims to localize and classify action instances by learning with only video-level category label. Most works focus on utilizing pre-extracted features for both RGB and optical flow modalities and learning with the multiple-instance learning strategy. Wang \etal \cite{wang2017untrimmednets} learn attention weights for each snippet and then threshold the attention weights for action proposals, which is extended by some other works~\cite{nguyen2018weakly,shi2020weakly,luo2020weakly} to achieve better localization. 
Several works~\cite{paul2018w,gong2020learning,islam2020weakly,min2020adversarial,zhang2021cola} propose a co-action similarity learning to enclose the video-level features between the video pairs with common classes.  Some works \cite{liu2019completeness,lee2020background,islam2021hybrid,nguyen2019weakly} seek to explicitly model the background activity for better foreground-background separation. Some works pursue the modality-wise consensus \cite{zhai2020two,yang2021uncertainty,narayan2021d2,ji2021weakly} for consistent predictions, which is extended by Hong \etal \cite{hong2021cross} for pre-extracted features re-calibration via a cross-modal consensus mechanism. Besides, Lee \etal \cite{lee2021weakly} model the background uncertainty on feature magnitude, while Huang \etal \cite{huang2021foreground} explore foreground-action consistency.

Though the works mentioned above make great progress, their performances are still inferior when compared with fully-supervised temporal action localization methods. One of the key factors for this result is the lack of snippet-level supervision signals. 
To address this issue, some methods~\cite{lee2021weakly,huang2022multi,zhai2020two} assign pseudo labels on class-agnostic Temporal Attention Mechanism (TAM) ~\cite{islam2021hybrid,shi2020weakly,li2022forcing} or class-specific Temporal Class Activation Sequence (T-CAM)~\cite{liu2019completeness,paul2018w,wang2017untrimmednets} by a preset threshold. And some methods \cite{luo2020weakly,pardo2021refineloc,he2022asm} assign pseudo labels based on predicted action proposals. All of them provide snippet-level supervision and achieve performance gains. Nevertheless, there is a common problem with these pseudo-label generation methods. That is, assigning category labels directly to snippets is prone to introduce noise due to model misclassification, which impairs the performance of localization. 
In contrast, we adopt an ideology of exclusion induction to progressively exclude categories that the snippets should not belong to based on different confidence levels. 

\noindent\textbf{Complementary Learning} trains a model using complementary category labels that the pattern does not belong to. It was first proposed by Ishida \etal~\cite{ishida2017learning} to reduce the cost of data collection for multi-classification tasks, which uses complementary labels to provide valid information for limited data. Then, subsequent methods explore how to effectively use complementary labels~\cite{ishida2019complementary,chou2020unbiased,yu2018learning,feng2020learning,gao2021discriminative,wang2021learning}. In addition, some other methods adopt the complementary labels to alleviate the noisy label problem~\cite{han2020sigua,kim2019nlnl,kim2021joint,xia2020part}. Different from the above methods, our proposed Progressive Complementary Learning is specifically designed for weakly-supervised temporal action localization. In our work, we consider two task-oriented properties as prior knowledge for progressive category exclusion, \ie, actions and the background cannot co-exist in one snippet and action instances of the same category have varied durations.


\section{Progressive Complementary Learning}
\label{sec:method}
In this section, we elaborate on our Progressive Complementary Learning (ProCL), which gradually enhances the snippet-level supervision from a category exclusion perspective.

\subsection{Problem Formulation and Model Overview}
\noindent\textbf{Problem Formulation.} 
Weakly-Supervised Temporal Action Localization (WSTAL) assumes that there are $N$ untrimmed videos denoted by $\{\mathcal{V}^{(i)}, Y^{(i)}\}_{i=1}^{N}$, where $Y^{(i)}=[y_1^{(i)},...,y_C^{(i)}]$ is the video-level category labels of the video $\mathcal{V}^{(i)}$. Specifically, $C$ denotes the number of action categories, and $y_j^{(i)}=1$ means the video $\mathcal{V}^{(i)}$ contains at least one action instance of the $j$-th category, while $y_j^{(i)}=0$ if the video $\mathcal{V}^{(i)}$ does not contain any action instance of the $j$-th category. In what follows, the superscript of all symbols will be omitted for convenience, \eg, $\mathcal{V}^{(i)}$ will be replaced by $\mathcal{V}$. WSTAL aims to train a model that can predict a set of action proposals during the inference phase and denotes them in the form of $\{(t_s,t_e,\psi,c)\}$, where $t_s$ and $t_e$ represent the start and end time, $c$ is the predicted action category, and $\psi$ represents the confidence score.  

\noindent\textbf{Overview.} \Cref{fig:two} illustrates the overall framework of our method. Given the features of the video $\mathcal{V}$, we sample a sequence of $T$ snippets uniformly in temporal dimension to obtain the original features $\mathbf{X}\in \mathbb{R}^{T\times D}$. The features $\mathbf{X}$ are then fed into the model to obtain the class activation sequence and the temporal attention scores for Multi-Instance Learning loss (\Cref{sec:mil}). Subsequently, based on the snippet-level supervision indicated by the video-level labels, we adopt Complementary Learning loss to exclude categories that all snippets in the video surely do not belong to (\Cref{sec:cl}). Then, to further enhance the snippet-level supervision, we propose Pseudo Complementary Labeling for snippets of less ambiguity to exclude more categories.  Then, the Foreground Background Discrimination loss is adopted to disambiguate the ambiguous snippets for coarsely enhancing the snippet-level supervision. (\Cref{sec:pcl}). Moreover, we utilize the collaborative information of multi-scale snippet sequences to further improve the quality of the snippet-level supervision (\Cref{sec:mpc}).

\subsection{Multiple-Instance Learning for WSTAL}
\label{sec:mil}
Following the previous WSTAL works \cite{lee2020background,islam2021hybrid,hong2021cross,li2022forcing}, we firstly introduce a background category and build a temporal attention branch for background suppression. Then the multi-instance learning loss is adopted for video-level learning. Specifically, we encode features $\mathbf{X}$ using several layers of temporal convolution to capture the temporal relationships between snippets and generate the class activation sequence denoted as $\mathcal{S}=\{s_{t,c}\}^{T,C+1}\in \mathbb{R}^{T\times (C+1)}$ through the classification head, where $s_{t,c}$ represents the activation score of the $c$-th category for the $t$-th snippet. And the background category is denoted as the $C+1$-th category for each snippet. Correspondingly, we extend $Y$ to $\widetilde{Y}=[y_1,...,y_C,1]\in\mathbb{R}^{C+1}$, which represents the existence of background context. 

Simultaneously, we input $\mathbf{X}$ into the attention head to get class-agnostic attention scores $\mathcal{A}=\{a_t\}^{T}\in\mathbb{R}^T$, and $a_t$ denotes the probability that the $t$-th snippet is an action. Next, we can obtain the background-free class activate sequence denoted as $\hat{\mathcal{S}}=\{\hat{s}_{t,c}\}^{T,C+1}\in\mathbb{R}^{T\times(C+1)}$ by suppressing the background context via $\mathcal{A}$, where $\hat{s}_{c,t}=s_{c,t}\cdot a_t$. Similarly, we can obtain background-free video-level labels $\hat{Y}=[y_1,...,y_C,0]\in \mathbb{R}^{C+1}$, which represents the inexistence of background context. 

Overall, the video-level classification loss is derived from Multi-Instance Learning (MIL) loss as follows, 
\begin{equation}
    L_{MIL} = L_{CE}(\widetilde{Y},\Phi(\Gamma(S)))+L_{CE}(\hat{Y},\Phi(\Gamma(\hat{S}))),
  \label{eq:mil}
\end{equation}
where $L_{CE}$ is the categorical cross-entropy error function, $\Phi{(\cdot)}$ represents the category-wise softmax function, and $\Gamma(\cdot)$ is the temporal Top-K pooling following the previous methods~\cite{chen2022dual,lee2020background,hong2021cross} to obtain the video-level prediction scores $\Gamma(\mathcal{S})\in \mathbb{R}^{C+1}$.

\subsection{Complementary Learning}
\label{sec:cl}

By analyzing the WSTAL problem, we note a fact that the video-level labels precisely indicate the categories that all snippets surely do not belong to, and we term these categories \textit{deterministic complementary categories}.
Therefore, we propose the Complementary Learning loss to exclude these deterministic complementary categories for all snippets in a video.

Specifically, the snippet-level classification scores $\mathcal{P}=\{p_{t,c}\}^{T,C+1}\in R^{T\times (C+1)}$ are first obtained by performing the category-wise softmax function on class activation sequence $\mathcal{S}$, and the Complementary Learning (CL) loss is formulated as follows: 
\begin{equation}
    L_{CL}=\frac{1}{T}\sum_{t=1}^{T}\sum_{c=1}^{C+1}-(1-y_{t,c})\log{(1-p_{t,c})},
  \label{eq:cl}
\end{equation}
where $y_{t,c}$ is the expansion of the video-level label $\widetilde{Y}_C$ in the temporal dimension, \ie, $y_{t,c}=\widetilde{Y}_C$. 

Through the Complementary Learning loss, we suppress the activation of the deterministic complementary categories for all snippets. Also, from another perspective, $L_{CL}$ promotes the activation of the ground truth categories in the video.
Although the pseudo-label-based methods also played a similar role, the noise problem of directly assigning pseudo labels deteriorated the performance of localization. In contrast, the deterministic complementary categories obtained from the video-level categories are free of noise and provide precise snippet-level supervision. 

\subsection{Pseudo Complementary Learning}
\label{sec:pcl}
Although above we have excluded the categories that all snippets surely do not belong to, there are still multiple categories remaining for the multi-label temporal action localization task.
To further enhance the snippet-level supervision, we attempt to exclude some non-existent categories from those video-level categories for the snippets of less ambiguity. Intuitively, it is easier to exclude some categories than exactly predict the (usually) one ground-truth category (i.e., excluding all other categories).

Therefore, we propose Pseudo Complementary Learning to further exclude some categories for snippets in the video. It consists of three parts, namely identifying the snippets of less ambiguity, excluding more categories for snippets of less ambiguity, and disambiguation for ambiguous snippets. 

\noindent\textbf{Ambiguity Identification by Information Entropy}. The ambiguous snippets are identified by the category information entropy of the class activation scores $\mathcal{S}$. Intuitively, the higher the category information entropy, the more likely the snippet is to be ambiguous. Specifically, we denote the indicator $\mathrm{F}=\{f_t\}^T \in \mathbb{R}^T$ of ambiguous snippets as follows: 
\begin{equation}
    f_t =  \left\{\begin{matrix}
     1,& \hbar_t<\theta, \\ 
     0, & \text{otherwise},
    \end{matrix}\right.
  \label{eq:identify}
\end{equation}
where $\theta$ is a preset threshold to estimate the 
ambiguity of the $t$-th snippet, and $\hbar_t$ represents the category information entropy of $t$-th snippet computed as follows: 
\begin{equation}
    \hbar_t  = -s_{t,\text{fg}}\cdot\log(s_{t,\text{fg}})-s_{t,\text{bg}}\cdot \log(s_{t,\text{bg}})
  \label{eq:five}
\end{equation}
\begin{equation}
    s_{t,\text{fg}} = \sum_{c\in \mathcal{G}\wedge c\ne\text{bg}}s_{t,c}  \ ,
  \label{eq:info_entropy}
\end{equation}
where $\mathcal{G}\in \mathbb{R}^{\mathbf{G}}$ is the category set except deterministic complementary categories, ``$\text{fg}$" denotes the set of foreground action categories and ``$\text{bg}$" denotes background category. 

\noindent\textbf{Background-aware Pseudo Complementary Labeling}. 
For the snippets of less ambiguity, \ie, $f_t=0$, we assign the snippet-level pseudo complementary labels $\mathcal{R}=\{r_{t,c}\}^{T,C+1}\in \mathbb{R}^{T\times (C+1)}$ to them. The purpose of this is to exclude the categories that snippets likely do not belong to, which we term \textit{pseudo complementary categories}. The process is formulated as follows: 
\begin{equation}
    r_{t,c} =  \left\{\begin{matrix}
     0,& \text{if}~\forall{c}\in \mathcal{G}, ~f_t=0 ~\text{and}~ s_{t,c}<\mu_t, \\ 
     1, & \text{otherwise},
    \end{matrix}\right.
  \label{eq:plg}
\end{equation}
where $\mu_t$ represents the mean value of activation scores in $t$-th snippet for the categories contained in $\mathcal{G}$. Considering the property that the foreground actions and background cannot co-exist in the same snippet, we further exclude categories with conflict. Specifically, after executing \cref{eq:plg} to exclude the pseudo complementary categories, if a snippet still has the background and at least one foreground action remaining, we exclude the one with a lower prediction score. Moreover, it is worth noting that the foreground is a set of action categories, and when the background is excluded, multiple action categories may be left. 

Based on the pseudo complementary labels $\mathcal{R}\in \mathbb{R}^{T\times (C+1)}$ obtained above and the snippet-level classification scores $\mathcal{P}\in \mathbb{R}^{T\times (C+1)}$, the Pseudo Complementary Learning (PCL) loss is derived below, 
\begin{equation}
    L_{PCL} = \frac{1}{N}\sum_{t=1}^{T}{\mathbb{I}(f_t=0)} \sum_{c\in \mathcal{G}} -(1-r_{t,c})\log{(1-p_{t,c})},
  \label{eq:eight}
\end{equation}
where $N=T-\sum_{t=1}^{T}f_t$ represents the amount of the snippet with less ambiguity, and $\mathbb{I}(f_t=0)$ means the ambiguous snippets are not involved in the $L_{PCL}$. 

\noindent\textbf{Disambiguation between Foreground and Background.} For the remaining ambiguous snippets, it is difficult to use Pseudo Complementary Labeling because the model is not confident about the prediction of these snippets. Therefore, we propose a Foreground Background Discrimination loss to reduce the ambiguity of the ambiguous snippets, which coarsely provides further snippet-level supervision for distinguishing foreground from background. 

Specifically, for the ambiguous snippets with $f_t=1$, we use the class-agnostic attention scores $\mathcal{A}$ generated by the attention branch and the background category activation scores $\mathcal{S}_{bg}=\mathcal{S}_{C+1}$ in class activation sequence for mutual learning. As \Cref{fig:two} shown, background attention scores can be obtained via $\mathcal{B}=(1-\mathcal{A})\in\mathbb{R}^T$, and denoted as $b_t=1-a_t$ for $t$-th snippet. Meanwhile, considering that the background prediction of different modules (\ie, $\mathcal{S}_{bg}$ and $\mathcal{B}$) should be consistent, we propose the Foreground-Background Discrimination (FBD) loss to disambiguate the ambiguous snippets, 
which is formulated as follows: 
\begin{equation}
    \begin{aligned}
    L_{FBD} &= \frac{1}{N}\sum_{t=1}^T \mathbb{I}(f_t=1)  E(b_t,p_{t,\text{bg}}) L_{BCE
    }(b_t,p_{t,\text{bg}})\\
        & + \frac{1}{N}\sum_{t=1}^T \mathbb{I}(f_t=1)  E(p_{t,\text{bg}},b_t) L_{BCE}(p_{t,\text{bg}},b_t), 
    \end{aligned}
   \label{eq:nine}
\end{equation}
where $L_{BCE}$ is a binary cross-entropy function, and $p_{t,\text{bg}}$ is the classification score of the background category of $t$-th snippet in snippet-level classification scores $P$. Besides, $\mathbb{I}(f_t=1)$ indicates that only ambiguous snippets are involved in the calculation as well as $N=T-\sum_{t=1}^T f_t$, and $E(\cdot)$ is a weight term based on Kullback-Leibler divergence as follows: 
\begin{equation}
   E(L,J)=\exp{(-l\cdot\log{(\frac{l}{j})})}. 
   \label{eq:ten}
\end{equation}

\subsection{Multi-Scale Complementary Labeling}
\label{sec:mpc}
All of the above are based on single-scale snippet sequence, \ie, 
sampling a sequence of $T$ snippets uniformly from a video. However, in real-world videos, action instances of the same category could be performed within different temporal durations. Taking ``Diving" as an example, it is obvious that jumping from platforms of different heights require various durations, but they are all ``Diving" actions. Based on the above observation, we additionally sample two snippet sequences to further improve the quality of the snippet-level supervision.

In detail, for densely sampled snippets $\mathbf{X}^{D}\in \mathbf{R}^{2T\times D}$ and sparsely sampled snippets $\mathbf{X}^{S}\in \mathbb{R}^{\frac{T}{2}\times D}$, we feed them into the feature extractor and classification head with shared parameters to obtain the corresponding class activation sequence $\mathcal{S}^D\in \mathbb{R}^{2T\times (C+1)}$ and $\mathcal{S}^S\in \mathbb{R}^{\frac{T}{2}\times (C+1)}$ respectively. Then they are re-scaled to the sequence of $T$ snippets and calculate the mean value and variance of the corresponding position with original class activation sequence $\mathcal{S}$ to obtain $\mathcal{S}^{\mu}\in \mathbb{R}^{T\times (C+1)}$ and $\mathcal{S}^{\sigma}\in \mathbb{R}^{T\times (C+1)}$. Here, $\mathcal{S}^\mu$ represents the average class activation sequence that incorporates multi-scale information, and $\mathcal{S^\sigma}$ denotes the inconsistency between sequences of different scales. Subsequently, we assign the pseudo complementary labels $\ddot{\mathcal{R}}$ and the indicator $\ddot{\mathbf{F}}$ of ambiguous snippets under multi-scale information using the same approach as \Cref{sec:pcl}. Therefore, the Multi-scale Pseudo Complementary Learning (MPCL) loss is derived as follows:
\begin{gather}
        L_{MPCL} = \frac{1}{N}
        \sum_{t=1}^{T}
        \mathbb{I}(\ddot{f}_t=0)D(\ddot{\mathcal{R}}_t,\mathcal{S}^\sigma_t,\mathcal{S}^{\mu}_t),
        \\D(L,U,J) = -\frac{1}{|\mathcal{G}|}\sum_{c\in \mathcal{G}}\exp(-u_c)(1-l_{c})log(1-j_{c}),
    \label{eq:ten}
\end{gather}
where $\mathbb{I}(\ddot{f}_t=0)$ means the ambiguous snippets are not involved in $L_{MPCL}$, and $\exp(\cdot)$ represents a weight term based on $\mathcal{S}^\sigma$, \ie, the larger variance $\mathcal{S^{\sigma}}$ of multi-scale outputs, the more likely there is a misjudgment and the weight should be reduced. 

\subsection{Overall Training Objective and Inference}
Overall, we combine all the loss functions mentioned above to obtain the final objective function of our model:
\begin{equation}
    L=L_{MIL}+L_{CL}+L_{MPCL}+L_{FBD}.
  \label{eq:thirteen}
\end{equation}

In the inference stage, following the standard process~\cite{chen2022dual,islam2021hybrid}, we first generate $\mathcal{A}$ and $\hat{\mathcal{S}}$ for the given video. Then the video-level classification scores are obtained based on $\hat{\mathcal{S}}$ to determine which action categories exist in the video by thresholding $\rho$. Next, class-agnostic action proposals are obtained from $\mathcal{A}$ via multiple thresholds. For the set of candidate action proposals $\{(t_s,t_e,\phi,c)\}$, we use non-maximal suppression to remove redundant proposals. 

\section{Experiments}
\label{sec:exp}
\subsection{Datasets and Evaluation Metrics}
We evaluate our method on two public benchmark datasets, \ie THUMOS14 \cite{THUMOS14} and ActivityNet1.3 \cite{caba2015activitynet}, for temporal action localization.

\noindent\textbf{THUMOS14} dataset contains 200 validation videos and 213 test videos of 20 action classes. It is a challenging benchmark, whose videos have diverse durations and frequent action instances. We use the validation videos as the training set and the test videos as the test set. 

\noindent\textbf{ActivityNet1.3} is a large dataset that covers 200 action categories, with a training set of 10,024 videos and a validation set of 4,926 videos. We use the training set and validation set for training and test, respectively.

\noindent\textbf{Evaluation metric.} We evaluate our method with mean average precision (mAP) under multiple of union (IoU) thresholds, which are the standard evaluation metrics for temporal action localization. We use the officially released evaluation method~\cite{caba2015activitynet} to measure our results.

\begin{table*}[t] 
	\centering
    \vspace{-5mm}
     \centering\caption{Comparisons with the state-of-the-art methods on the THUMOS14 dataset. AVG 0.1:0.5 and AVG 0.1:0.7 are the average mAP computed under thresholds [0.1:0.1:0.5] and [0.1:0.1:0.7], separately. $\dagger$ denotes the method adopting pseudo labels. And $\ddag$ means using additional information, e.g. frequency or pose. }

    \scalebox{0.8}{
		\begin{tabular}{c|c|c||ccccccc|c|c}
		    \hline
		    \multirow{2}{*}{Supervision}  & \multirow{2}{*}{Method} &  \multirow{2}{*}{Publication}& \multicolumn{7}{c|}{mAP@IoU (\%)} & AVG & AVG\\
		    \cline{4-10}
		     & & & 0.1 & 0.2 & 0.3 & 0.4 & 0.5 & 0.6 & 0.7 & 0.1:0.5 & 0.1:0.7\\ 
		    \hline
		    \hline
		    \multirow{5}{*}{Full} 
		     & SSN \cite{zhao2017temporal} &ICCV 2017 &  60.3 & 56.2 & 50.6 & 40.8 & 29.1	& - & - &47.4& -\\
		     & BSN \cite{lin2018bsn}&ECCV 2018&  - & - & 53.5 & 45.0 & 36.9 & 28.4 & 20.0  & - & -\\
		     & BMN \cite{lin2019bmn} &ICCV 2019 &  - & - & 56.0 & 47.4& 38.8 & 29.7 & 20.5 &-&-\\ 
		    & P-GCN \cite{zeng2019graph}&ICCV 2019 & 69.5 & 67.8 & 63.6 & 57.8 & 49.1 & - & - &61.6&-\\
            & G-TAD \cite{xu2020g}&CVPR 2020&-&-&66.4&60.4&51.6&37.6&22.9&-&-\\
            \hline

            \multirow{4}{*}{Weak $^{\ddag}$} 
            & STAR \cite{xu2019segregated}&AAAI 2019 & 68.8	& 60.0 & 48.7 & 34.7 & 23.0	& - & -  &47.0&-\\
            & 3C-NET \cite{narayan20193c}&ICCV 2019 & 59.1 & 53.5 & 44.2 & 34.1 & 26.6 & - & 8.1 &43.5&-\\
            & PreTrimNet \cite{zhang2020multi}&AAAI 2020 & 57.5	& 50.7 & 41.4 & 32.1 & 23.1	& 14.2 & 7.7 & 41.0 &32.4 \\
            & SF-Net \cite{ma2020sf} &ECCV 2020 &71.0 & 63.4 & 53.2 & 40.7 & 29.3 & 18.4 & 9.6 & 51.5 &40.8\\

            
            \hline
            \multirow{22}{*}{Weak} 
            & BaS-Net \cite{lee2020background} &AAAI 2020 &  58.2 & 52.3 & 44.6 & 36.0 & 27.0 & 18.6 & 10.4 & 43.6 &35.3\\
            & $\dagger$ EM-MIL \cite{luo2020weakly}&ECCV 2020 &  59.1 & 52.7 & 45.5 & 36.8	& 30.5 & 22.7 & 16.4&44.9&37.7\\
            & A2CL-PT \cite{min2020adversarial} &ECCV 2020 & 61.2 & 56.1& 48.1 & 39.0 & 30.1 & 19.2 & 10.6 &46.9&37.8 \\
            & $\dagger$ TSCN \cite{zhai2020two}&ECCV 2020 &  63.4 & 57.6 & 47.8 & 37.7 & 28.7 & 19.4 & 10.2&  47.0&37.8\\
            & $\dagger$ UM \cite{lee2021weakly} &AAAI 2021 & 67.5	& 61.2 & 52.3 & 43.4 & 33.7	& 22.9 & 12.1 & 51.6 &41.9\\
            & CoLA \cite{zhang2021cola} &CVPR 2021 &  66.2&59.5&51.5&41.9&32.2&22.0&13.1&50.3&40.9\\
            & AUMN \cite{luo2021action} &CVPR 2021& 66.2&61.9&54.9&44.4&33.3&20.5&9.0&52.1&41.5\\
            & FAC-Net \cite{huang2021foreground} &ICCV 2021 & 67.6 & 62.1 & 52.6 & 44.3 & 33.4 & 22.5 & 12.7 & 52.0&42.2 \\
            & D2-Net \cite{narayan2021d2} &ICCV 2021 &  65.7& 60.2 & 52.3 & 43.4 & 36.0 &- & - & 51.5&-  \\
            & $\dagger$ UGCT \cite{yang2021uncertainty} &CVPR 2021&  69.2&62.9 &55.5&46.5&35.9&23.8&11.4&54.0&43.6\\
            & HAM-Net \cite{islam2021hybrid} &AAAI 2021 &  65.4	& 59.0 & 50.3 & 41.1 & 31.0	& 20.7 & 11.1& 49.4 &39.8\\
            & $\dagger$ RefineLoc \cite{pardo2021refineloc}&WACV 2021&-&-&40.8 &32.7&23.1&13.3&5.3&-&-\\
            & CSCL \cite{ji2021weakly} &MM 2021& 68.0 & 61.8 & 52.7 & 43.3 & 33.4 & 21.8 & 12.3 & 51.8 &41.9\\
            & CO$_2$-Net \cite{hong2021cross} &MM 2021&  70.1 & 63.6 & 54.5 & 45.7 & 38.3 & 26.4 & 13.4 & 54.4 &44.6\\
            & ACGNet \cite{yang2022acgnet} &AAAI 2022&  68.1 & 62.6 & 53.1 & 44.6 & 34.7 & 22.6 & 12.0 &52.6&42.5 \\
            & FTCL \cite{gao2022fine}&CVPR 2022 & 69.6& 63.4&55.2&45.2&35.6&23.7&12.2&53.8&43.6\\
            & DCC \cite{li2022exploring} &CVPR 2022&69 & 63.8 &55.9&45.9&35.7&24.3&13.7&54.1&44.0\\
            & $\dagger$ Huang \etal \cite{huang2022weakly} &CVPR 2022&71.3&65.3&55.8&47.5&38.2&25.4&12.5&55.6&45.1\\
            & $\dagger$ ASM-Loc \cite{he2022asm} &CVPR 2022& 71.2& 65.5&57.1&46.8&36.6&25.2&13.4&55.4&45.1\\
            & DELU \cite{chen2022dual} &ECCV 2022&71.5&66.2&56.5&47.7&40.5&27.2&15.3&\underline{56.5}&\underline{46.4}\\
            & DGCNN \cite{shi2022dynamic}&MM 2022& 66.3& 59.9&52.3&43.2&32.8&22.1&13.1&50.9&41.3\\
            & Li \etal \cite{li2022forcing} &MM 2022 & 69.7 &64.5&58.1&49.9&39.6&27.3&14.2&56.3&46.1\\
            
            \cline{2-12}
            & \textbf{ProCL} & \textbf{Ours} & 73.3 & 68.2 &59.6 & 49.3 & 40.5 & 27.8& 14.9 & \textbf{58.2} & \textbf{47.7}\\
            \hline
		\end{tabular}
		}
      \label{tab:comp_thumos}

    \vspace{-2mm}
\end{table*}%

\subsection{Implementation Details}
We apply I3D \cite{carreira2017quo} pre-trained on Kinetics-400 \cite{kay2017kinetics} to extract both RGB and optical flow features, which are concatenated into 2048-dimensional snippet-level features. Each snippet contains continuous non-overlapping 16 frames from the original untrimmed video.
In the training stage, we set $T=560$ and $T=90$ for THUMOS14 and ActivityNet1.3 respectively, while all snippets are taken during evaluation.
Adam optimizer with 0.001 weight decay rate and 0.00003 learning rate is used. 
The number of training iterations for THUMOS14 and ActivityNet1.3 is set to $20000$ and $30000$, $\gamma$ of top-k pooling that $k=T/\gamma$ is set to $7$ and $10$. 
Also, we perform the experiments multiple times with different seeds to report the mean values. The method is implemented in PyTorch \cite{paszke2019pytorch} and all experiments are performed on an NVIDIA GTX 1080Ti GPU. More details are available in \textit{Appendix}.

\begin{table}[th]
    \centering
    \caption{Comparisons with other methods on the ActivityNet1.3 dataset. AVG mAP means average mAP from IoU 0.5 to 0.95 with 0.05 increment. $\dagger$ denotes the method adopting pseudo labels. And $\ddag$ means using additional information, e.g. frequency or pose. }
    \resizebox{1.0\columnwidth}{!}{
        \begin{tabular}{c|c|c||ccc||c}
        \hline
        Supervision&Method &Publication& 0.5 & 0.75& 0.95 & AVG mAP\\
        \hline
        \hline
        \multirow{3}{*}{Full} 

        &TAL-Net \cite{chao2018rethinking}&CVPR 2018&  38.2 & 18.3 & 1.3 & 20.2\\
        & BSN \cite{lin2018bsn}&ECCV 2018& 46.5& 30.0 & 8.0 & 30.0\\
        &P-GCN \cite{zeng2019graph}&ICCV 2019& 42.9 & 28.1& 2.5& 27.0\\
        \hline
        
        \multirow{3}{*}{Weak$^{\ddag}$} 
        &TSRNet \cite{zhang2019learning}&AAAI 2019& 33.1 & 18.7 & 3.3 & 21.8\\
        &STAR \cite{xu2019segregated} &AAAI 2019 & 31.1 & 18.8 &4.7 & - \\
        &PreTrimNet \cite{zhang2020multi}&AAAI 2020& 34.8 & 20.9 & 5.3 & 22.5\\
        \hline
        
        \multirow{13}{*}{Weak} 
        &BaS-Net \cite{lee2020background}&AAAI 2020 & 34.5 & 22.5 & 4.9& 22.2\\
        &A2CL-PT \cite{min2020adversarial}&ECCV 2020 & 36.8 & 22.0 & 5.2 & 22.5\\
        &$\dagger$ TSCN \cite{zhai2020two} &ECCV 2020& 35.3 & 21.4 & 5.3 & 21.7 \\
        &ACSNet \cite{liu2021acsnet} &AAAI 2021& 36.3 & 24.2 & 5.8 & 23.9 \\
        &$\dagger$ UM \cite{lee2021weakly}&AAAI 2021 & 37.0 & 23.9 & 5.7 & 23.7\\
        &AUMN \cite{luo2021action}&CVPR 2021& 38.3 & 23.5 & 5.2 & 23.5 \\
        &$\dagger$ UGCT \cite{yang2021uncertainty}&CVPR  2021 & 39.1 & 22.4 & 5.8 & 23.8\\
        &FAC-Net \cite{huang2021foreground}&ICCV 2021 & 37.6 & 24.2 & 6.0 & 24.0 \\
        & FTCL \cite{gao2022fine}&CVPR 2022 & 40.0& 24.3&6.4 &24.8\\
        & DCC \cite{li2022exploring} &CVPR 2022&38.8& 24.2 & 5.7& 24.3 \\
        &$\dagger$  Huang \etal \cite{huang2022weakly} &CVPR 2022&40.6& 24.6& 5.9& 25.0 \\
        &$\dagger$ ASM-Loc \cite{he2022asm} &CVPR 2022&41.0 &24.9 &6.2 &\underline{25.1} \\
        & DGCNN \cite{shi2022dynamic}&MM 2022&37.2 &23.8 &5.8 &23.9 \\

        \cline{2-7}
        &\textbf{ProCL} & \textbf{Ours} &41.6 &26.1 & 6.0 & \textbf{26.1} \\
        \hline
        \end{tabular}
    }
    \centering

    \label{tab:comp_act}
    \vspace{-3mm}
\end{table}

\subsection{Comparison with State-of-the-Arts}
In \Cref{tab:comp_thumos}, we compare our proposed method with the state-of-the-art WSTAL methods on THUMOS14. We find that our method outperforms all previous methods in terms of AVG 0.1:0.5 and AVG 0.1:0.7. In particular, our approach outperforms all methods that adopt pseudo labeling to obtain snippet-level supervision, including both ways of assigning pseudo labels with preset thresholds \cite{lee2021weakly,zhai2020two,huang2022weakly} and assigning pseudo labels based on predicted action proposals \cite{he2022asm,pardo2021refineloc}. Also, our approach surpasses some methods~\cite{ma2020sf,zhang2020multi} that use additional information to provide more supervision. Moreover, the results of our method are even comparable with fully-supervised in terms of mAP@0.1 and mAP@0.2, and surpass SSN \cite{zhao2017temporal}. The results demonstrate the superior performance of our ProCL, which obtains relatively accurate snippet-level supervisory signals by progressive category exclusion. 

We also conduct experiments on the ActivityNet1.3 dataset and the results are reported in \Cref{tab:comp_act}. Consistent with THUMOS14, our method also achieves state-of-the-art performance, surpassing the latest methods. The consistent conclusions on both datasets demonstrate the effectiveness of our ProCL. We also conduct experiments on additional UNT features and also achieve state-of-the-art performance, see \textit{Appendix} for details.

\subsection{Ablation Studies}
\begin{table}[t]
    \centering
    \caption{Ablation studies on THUMOS14. AVG is the average mAP for IoU thresholds 0.1:0.7. Exp.1 is the baseline and Exp.7 is our full method. Exp.2 is the pseudo-label-based method following previous work~\cite{he2022asm}. }
        \resizebox{1\linewidth}{!}{
        \begin{tabular}{c|ccccc|c||c}
            \hline
             Exp&$\mathcal{L}_{MIL}$ & $\mathcal{L}_{CL}$ &$\mathcal{L}_{PCL}$ & $\mathcal{L}_{FBD}$ & $\mathcal{L}_{MPCL}$ &$\mathcal{L}_{PL}$&  AVG \\ \hline

            1&\checkmark  &  &  & & & & 41.7 \\
            \hline
            \hline
            2&\checkmark & & & & &\checkmark &44.0 \\ 
            \hline
            \hline
            3 &\checkmark &\checkmark & & & & &45.9 \\
            4&\checkmark & \checkmark&\checkmark & & & &46.5 \\
            5&\checkmark &\checkmark& &\checkmark & & &46.4 \\
            6&\checkmark &\checkmark &\checkmark&\checkmark& & &47.3 \\
            7&\checkmark &\checkmark & & \checkmark&\checkmark & &47.7 \\
            \hline

        \end{tabular}
    }

    \label{tab:ablation}
    \vspace{-3mm}
\end{table}
\begin{figure}[t]
    \centering
    \vspace{-3mm}
    \includegraphics[width=\linewidth]{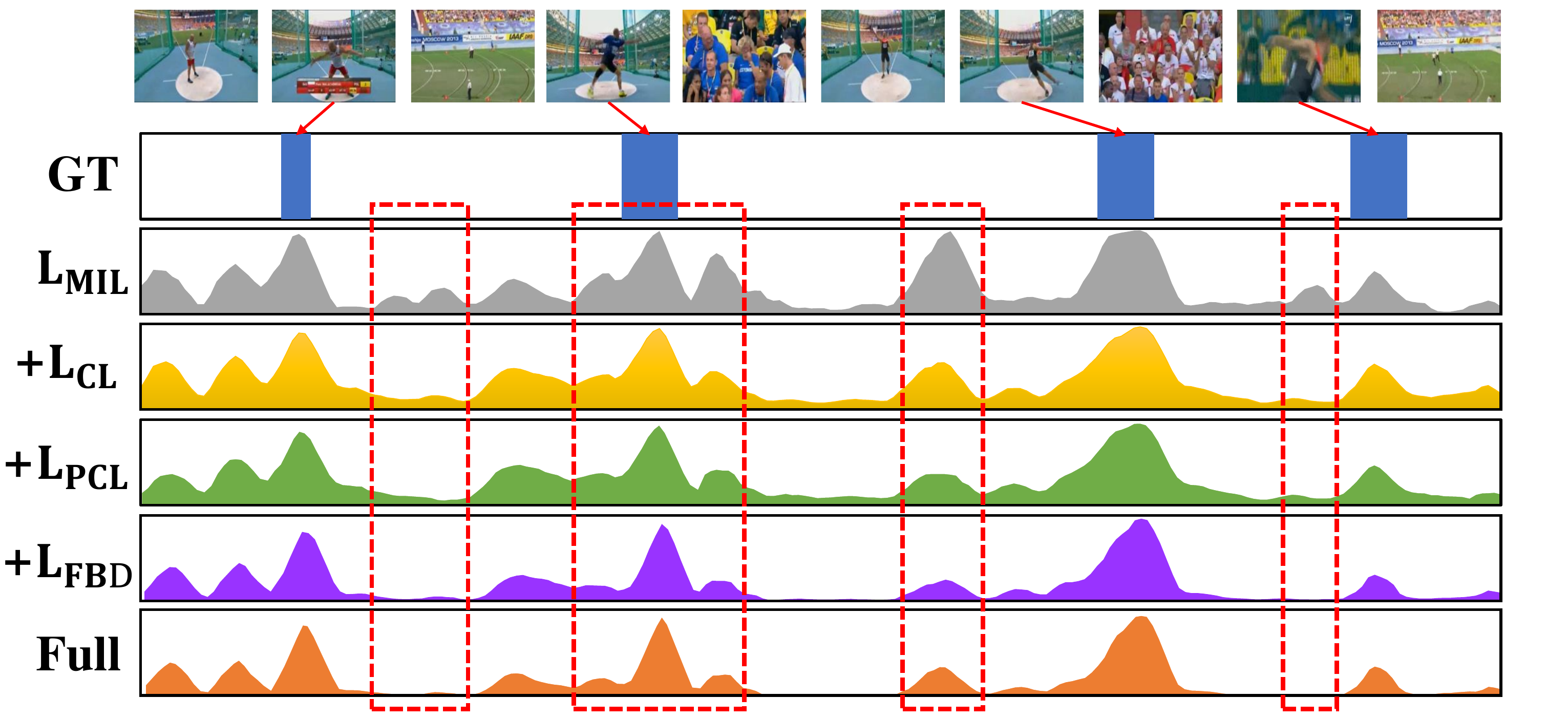}
    \caption{Visualization of ablation studies, where ``+" indicates a new component is added upon the previous experiment. Red dash boxes note areas of significant improvement.}
    \label{fig:ab}
    \vspace{-1mm}
\end{figure}
\noindent\textbf{Quantitative analysis of each component.} In \Cref{tab:ablation}, we show the quantitative analysis of the different components of our ProCL. We use ``AVG" for the performance metric, which is the average of mAP values for different IoU thresholds [0.1:0.1:0.7]. Exp.1 is the baseline only trained with $L_{MIL}$, and Exp.7 is our full method. By using the Complementary Learning loss $L_{CL}$, Exp.3 improves by 4.2\% compared with Exp.1, which attributes to accurately exploiting the snippet-level supervision indicated by video-level labels. Surprisingly, our $L_{CL}$ outperforms the previous pseudo-label-based approach $L_{PL}$ in Exp.2 by 1.9\%, which benefits from the fact that our $L_{CL}$ is noise-free and provides the appropriate snippet-level supervision for model training. Exp.4 shows 0.6\% performance improvement over Exp.3, which demonstrates that our Pseudo Complementary Learning further provides effective snippet-level supervision for the model. By introducing the Foreground Background Discrimination loss, Exp.6 achieve further performance gains by 0.8\%. Moreover, Exp.7 shows that the information fusion of multi-scale snippet sequences further improves the action localization performance. More analysis is given in the \textit{Appendix}.

\noindent\textbf{Qualitative analysis of each component.} In \Cref{fig:ab}, we visualize the qualitative results for the different components, where ``GT" represents the ground truth. Compared with the second row, the results of the third row show that the activation of some background snippets is obviously suppressed, which demonstrates that our deterministic complementary categories are able to improve the performance of action localization. Meanwhile, from the fourth row of the figure, we find that our pseudo complementary labels also show a positive effect on the localization results. Furthermore, consistent with the quantitative results, with the introduction of $L_{FBD}$ (the fifth row in the figure), the background snippets are further suppressed, demonstrating that our Foreground Background Discrimination loss has the ability to disambiguate ambiguous snippets.

\begin{figure}[t]
    \centering
    \vspace{-3mm}
    \includegraphics[width=0.90\linewidth]{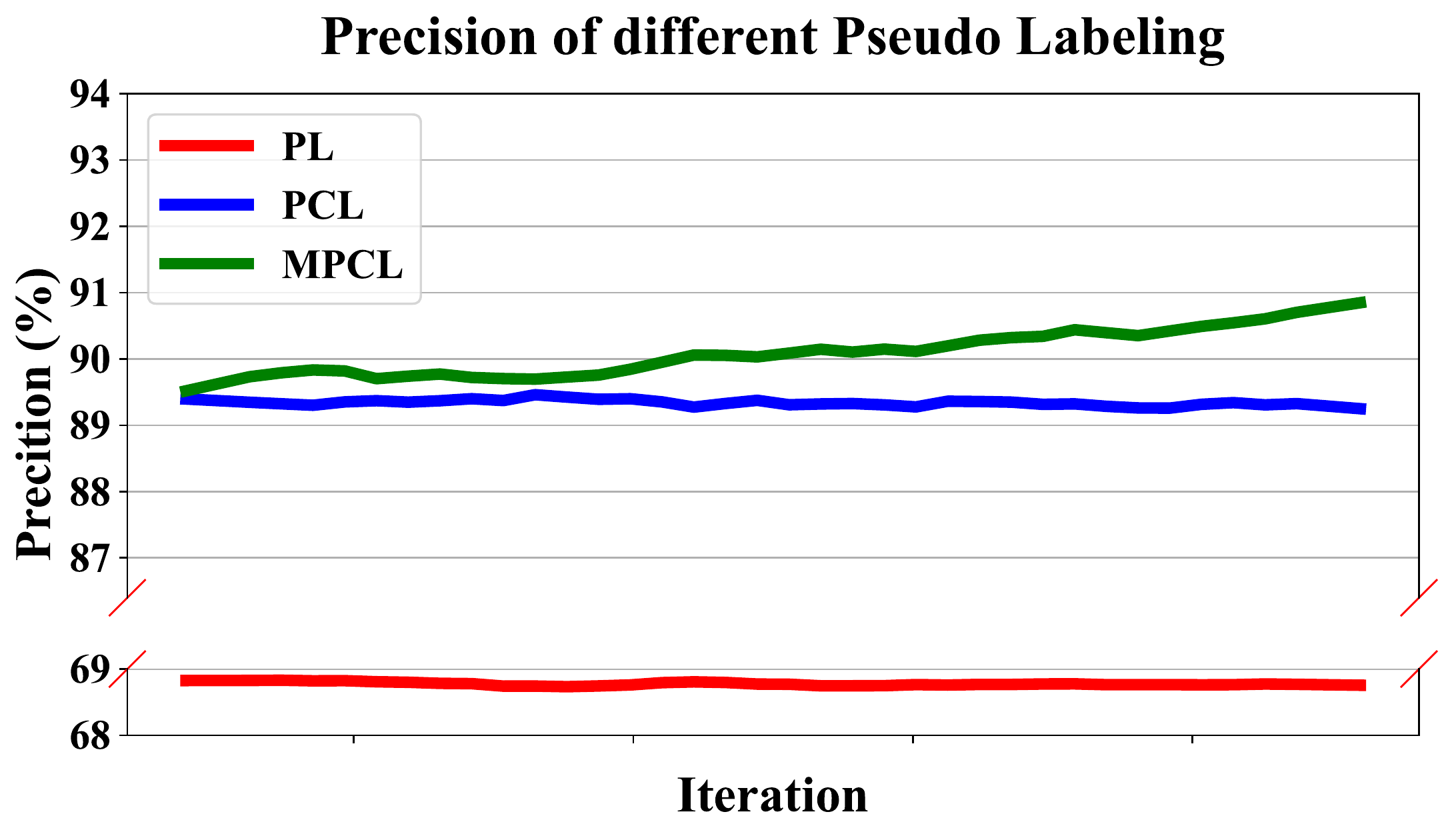}
    \vspace{-3mm}
    \caption{The label precision of different labeling methods, where \textbf{PL} denotes the pseudo labels obtained by following previous pseudo-label-based method~\cite{he2022asm}, \textbf{PCL} and \textbf{MPCL} denotes our pseudo complementary labels obtained from single-scale snippet sequence and multi-scale snippet sequences, respectively. }
    \label{fig:prec}
    \vspace{-1mm}
\end{figure}

\noindent\textbf{Label precision of different labeling methods.} 
In \Cref{fig:prec} we compare the precision of the labels that are assigned by the previous pseudo-label-based method and our pseudo complementary labeling. With just the single-scale snippet sequence, our method has a 20\% precision improvement. By comparing PCL and MPCL, it is found that the collaborative information of multi-scale snippet sequences indeed contributes to improving the quality of pseudo complementary labels. Not only that, but we also found that the precision of our pseudo complementary labels has improved with iterative refinement. This indicates that the pseudo complementary labels contribute to the predictions of the model and vice versa. These further demonstrate that our pseudo complementary labeling is superior to previous pseudo labeling methods, which alleviates the noise problem introduced by assigning category labels directly. 

\begin{figure}[t]
    \centering
    \includegraphics[width=\linewidth]{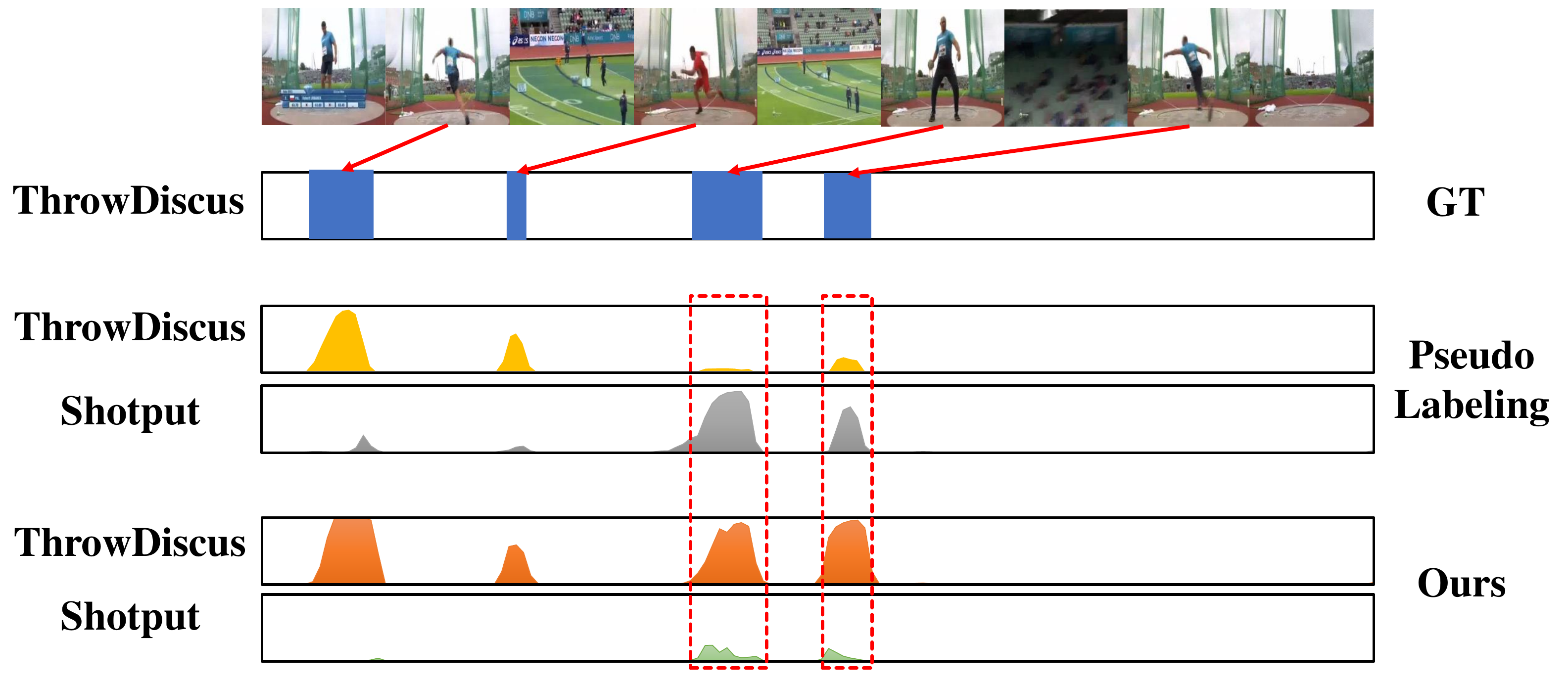}
    \caption{Visualization of category prediction scores of our and pseudo labeling. This figure shows a video containing only ``ThrowDiscus", and each row in the figure indicates the category prediction scores of the snippets. The red dashed boxes refer to the regions where the pseudo-label-based method misclassifies, but our method still classifies correctly.}
    \label{fig:pl_noisy}
    \vspace{-4mm}
\end{figure}

\noindent\textbf{Qualitative analysis by category prediction scores.} In \Cref{fig:pl_noisy}, we visualize the prediction scores of two similar actions (\ie, ``ThrowDiscus" and ``Shotput") in the category activation sequence $\mathcal{S}$. We find that the pseudo-label-based method incorrectly predicts an instance of ``ThrowDiscus'' as ``Shotput'' (\ie, denoted by red dash boxes), even though this video does not contain any instance of the ``Shotput".
In contrast, our method correctly distinguishes ``ThrowDiscus" from ``Shotput", as the activation of ``Shotput'' are suppressed. This result demonstrates that our ProCL can provide the correct snippet-level supervision information, which effectively suppresses the prediction of categories that snippets should not belong to.


\section{Conclusion}
In this work, we propose a \textit{Progressive Complementary Learning} (ProCL) method for weakly-supervised temporal action localization.
With the video-level labels, we first exclude the complementary categories that are surely not existent in the video.
And then, we exclude the pseudo complementary categories for the snippets with less ambiguity. Then, for ambiguous snippets where it is difficult to assign pseudo complementary labels directly, we disambiguate them to further obtain coarse snippet-level supervision. Furthermore, better pseudo complementary labels are obtained by employing collaborative information between multi-scale snippet sequences to provide higher quality snippet-level supervision. 
Experimental results demonstrate that our ProCL does provide effective snippet-level supervision to the model by progressively excluding categories that snippets should not belong to, and thus our ProCL makes significant improvements on weakly-supervised temporal action localization.

{\small

\bibliographystyle{ieee_fullname}
\bibliography{main}

\begin{thebibliography}{10}\itemsep=-1pt

\bibitem{caba2015activitynet}
Fabian Caba~Heilbron, Victor Escorcia, Bernard Ghanem, and Juan Carlos~Niebles.
\newblock {ActivityNet: A Large-Scale Video Benchmark for Human Activity
  Understanding}.
\newblock In {\em CVPR}, 2015.

\bibitem{carreira2017quo}
Joao Carreira and Andrew Zisserman.
\newblock {Quo Vadis, Action Recognition? A New Model and the Kinetics
  Dataset}.
\newblock In {\em CVPR}, 2017.

\bibitem{chao2018rethinking}
Yu-Wei Chao, Sudheendra Vijayanarasimhan, Bryan Seybold, David~A Ross, Jia
  Deng, and Rahul Sukthankar.
\newblock {Rethinking the Faster R-CNN Architecture for Temporal Action
  Localization}.
\newblock In {\em CVPR}, 2018.

\bibitem{chen2022dual}
Mengyuan Chen, Junyu Gao, Shicai Yang, and Changsheng Xu.
\newblock {Dual-Evidential Learning for Weakly-supervised Temporal Action
  Localization}.
\newblock In {\em ECCV}, 2022.

\bibitem{chou2020unbiased}
Yu-Ting Chou, Gang Niu, Hsuan-Tien Lin, and Masashi Sugiyama.
\newblock {Unbiased Risk Estimators Can Mislead: A Case Study of Learning with
  Complementary Labels}.
\newblock In {\em ICML}, 2020.

\bibitem{ciptadi2014movement}
Arridhana Ciptadi, Matthew~S Goodwin, and James~M Rehg.
\newblock {Movement Pattern Histogram for Action Recognition and Retrieval}.
\newblock In {\em ECCV}, 2014.

\bibitem{feng2021mist}
Jia-Chang Feng, Fa-Ting Hong, and Wei-Shi Zheng.
\newblock {MIST: Multiple Instance Self-Training Framework for Video Anomaly
  Detection}.
\newblock In {\em CVPR}, 2021.

\bibitem{feng2020learning}
Lei Feng, Takuo Kaneko, Bo Han, Gang Niu, Bo An, and Masashi Sugiyama.
\newblock {Learning with Multiple Complementary Labels}.
\newblock In {\em ICML}, 2020.

\bibitem{gao2022fine}
Junyu Gao, Mengyuan Chen, and Changsheng Xu.
\newblock {Fine-Grained Temporal Contrastive Learning for Weakly-Supervised
  Temporal Action Localization}.
\newblock In {\em CVPR}, 2022.

\bibitem{gao2017turn}
Jiyang Gao, Zhenheng Yang, Kan Chen, Chen Sun, and Ram Nevatia.
\newblock {TURN TAP: Temporal Unit Regression Network for Temporal Action
  Proposals}.
\newblock In {\em ICCV}, 2017.

\bibitem{gao2021discriminative}
Yi Gao and Min-Ling Zhang.
\newblock {Discriminative Complementary-Label Learning with Weighted Loss}.
\newblock In {\em ICML}, 2021.

\bibitem{gong2020learning}
Guoqiang Gong, Xinghan Wang, Yadong Mu, and Qi Tian.
\newblock {Learning Temporal Co-Attention Models for Unsupervised Video Action
  Localization}.
\newblock In {\em CVPR}, 2020.

\bibitem{han2020sigua}
Bo Han, Gang Niu, Xingrui Yu, Quanming Yao, Miao Xu, Ivor Tsang, and Masashi
  Sugiyama.
\newblock {SIGUA: Forgetting May Make Learning with Noisy Labels More Robust}.
\newblock In {\em ICML}, 2020.

\bibitem{he2022asm}
Bo He, Xitong Yang, Le Kang, Zhiyu Cheng, Xin Zhou, and Abhinav Shrivastava.
\newblock {ASM-Loc: Action-Aware Segment Modeling for Weakly-Supervised
  Temporal Action Localization}.
\newblock In {\em CVPR}, 2022.

\bibitem{hong2021cross}
Fa-Ting Hong, Jia-Chang Feng, Dan Xu, Ying Shan, and Wei-Shi Zheng.
\newblock {Cross-modal Consensus Network for Weakly Supervised Temporal Action
  Localization}.
\newblock In {\em ACM MM}, 2021.

\bibitem{hong2020mini}
Fa-Ting Hong, Xuanteng Huang, Wei-Hong Li, and Wei-Shi Zheng.
\newblock {MINI-Net: Multiple Instance Ranking Network for Video Highlight
  Detection}.
\newblock In {\em ECCV}, 2020.

\bibitem{huang2021foreground}
Linjiang Huang, Liang Wang, and Hongsheng Li.
\newblock {Foreground-Action Consistency Network for Weakly Supervised Temporal
  Action Localization}.
\newblock In {\em ICCV}, 2021.

\bibitem{huang2022multi}
Linjiang Huang, Liang Wang, and Hongsheng Li.
\newblock {Multi-Modality Self-Distillation for Weakly Supervised Temporal
  Action Localization}.
\newblock {\em TIP}, 2022.

\bibitem{huang2022weakly}
Linjiang Huang, Liang Wang, and Hongsheng Li.
\newblock {Weakly Supervised Temporal Action Localization via Representative
  Snippet Knowledge Propagation}.
\newblock In {\em CVPR}, 2022.

\bibitem{ishida2017learning}
Takashi Ishida, Gang Niu, Weihua Hu, and Masashi Sugiyama.
\newblock {Learning from Complementary Labels}.
\newblock {\em NeurIPS}, 2017.

\bibitem{ishida2019complementary}
Takashi Ishida, Gang Niu, Aditya Menon, and Masashi Sugiyama.
\newblock {Complementary-Label Learning for Arbitrary Losses and Models}.
\newblock In {\em ICML}, 2019.

\bibitem{islam2021hybrid}
Ashraful Islam, Chengjiang Long, and Richard Radke.
\newblock {A Hybrid Attention Mechanism for Weakly-Supervised Temporal Action
  Localization}.
\newblock In {\em AAAI}, 2021.

\bibitem{islam2020weakly}
Ashraful Islam and Richard Radke.
\newblock {Weakly Supervised Temporal Action Localization Using Deep Metric
  Learning}.
\newblock In {\em WACV}, 2020.

\bibitem{ji2021weakly}
Yuan Ji, Xu Jia, Huchuan Lu, and Xiang Ruan.
\newblock {Weakly-Supervised Temporal Action Localization via Cross-Stream
  Collaborative Learning}.
\newblock In {\em ACM MM}, 2021.

\bibitem{THUMOS14}
Y.-G. Jiang, J. Liu, A. Roshan~Zamir, G. Toderici, I. Laptev, M. Shah, and R.
  Sukthankar.
\newblock {THUMOS} challenge: Action recognition with a large number of
  classes.
\newblock \url{http://crcv.ucf.edu/THUMOS14/}, 2014.

\bibitem{kay2017kinetics}
Will Kay, Joao Carreira, Karen Simonyan, Brian Zhang, Chloe Hillier, Sudheendra
  Vijayanarasimhan, Fabio Viola, Tim Green, Trevor Back, Paul Natsev, et~al.
\newblock {The Kinetics Human Action Video Dataset}.
\newblock {\em arXiv}, 2017.

\bibitem{kim2019nlnl}
Youngdong Kim, Junho Yim, Juseung Yun, and Junmo Kim.
\newblock {NLNL: Negative Learning for Noisy Labels}.
\newblock In {\em ICCV}, 2019.

\bibitem{kim2021joint}
Youngdong Kim, Juseung Yun, Hyounguk Shon, and Junmo Kim.
\newblock {Joint Negative and Positive Learning for Noisy Labels}.
\newblock In {\em CVPR}, 2021.

\bibitem{lee2020background}
Pilhyeon Lee, Youngjung Uh, and Hyeran Byun.
\newblock {Background Suppression Network for Weakly-Supervised Temporal Action
  Localization}.
\newblock In {\em AAAI}, 2020.

\bibitem{lee2021weakly}
Pilhyeon Lee, Jinglu Wang, Yan Lu, and Hyeran Byun.
\newblock {Weakly-supervised Temporal Action Localization by Uncertainty
  Modeling}.
\newblock In {\em AAAI}, 2021.

\bibitem{li2022exploring}
Jingjing Li, Tianyu Yang, Wei Ji, Jue Wang, and Li Cheng.
\newblock {Exploring Denoised Cross-Video Contrast for Weakly-Supervised
  Temporal Action Localization}.
\newblock In {\em CVPR}, 2022.

\bibitem{li2022forcing}
Ziqiang Li, Yongxin Ge, Jiaruo Yu, and Zhongming Chen.
\newblock {Forcing the Whole Video as Background: An Adversarial Learning
  Strategy for Weakly Temporal Action Localization}.
\newblock In {\em ACM MM}, 2022.

\bibitem{lin2019bmn}
Tianwei Lin, Xiao Liu, Xin Li, Errui Ding, and Shilei Wen.
\newblock {BMN: Boundary-Matching Network for Temporal Action Proposal
  Generation}.
\newblock In {\em ICCV}, 2019.

\bibitem{lin2017single}
Tianwei Lin, Xu Zhao, and Zheng Shou.
\newblock {Single Shot Temporal Action Detection}.
\newblock In {\em ACM MM}, 2017.

\bibitem{lin2018bsn}
Tianwei Lin, Xu Zhao, Haisheng Su, Chongjing Wang, and Ming Yang.
\newblock {BSN: Boundary Sensitive Network for Temporal Action Proposal
  Generation}.
\newblock In {\em ECCV}, 2018.

\bibitem{liu2019completeness}
Daochang Liu, Tingting Jiang, and Yizhou Wang.
\newblock {Completeness Modeling and Context Separation for Weakly Supervised
  Temporal Action Localization}.
\newblock In {\em CVPR}, 2019.

\bibitem{liu2019learning}
Xingyu Liu, Joon-Young Lee, and Hailin Jin.
\newblock {Learning Video Representations From Correspondence Proposals}.
\newblock In {\em CVPR}, 2019.

\bibitem{liu2021acsnet}
Ziyi Liu, Le Wang, Qilin Zhang, Wei Tang, Junsong Yuan, Nanning Zheng, and Gang
  Hua.
\newblock {ACSNet: Action-Context Separation Network for Weakly Supervised
  Temporal Action Localization}.
\newblock In {\em AAAI}, 2021.

\bibitem{long2019gaussian}
Fuchen Long, Ting Yao, Zhaofan Qiu, Xinmei Tian, Jiebo Luo, and Tao Mei.
\newblock {Gaussian Temporal Awareness Networks for Action Localization}.
\newblock In {\em CVPR}, 2019.

\bibitem{luo2021action}
Wang Luo, Tianzhu Zhang, Wenfei Yang, Jingen Liu, Tao Mei, Feng Wu, and
  Yongdong Zhang.
\newblock {Action Unit Memory Network for Weakly Supervised Temporal Action
  Localization}.
\newblock In {\em CVPR}, 2021.

\bibitem{luo2020weakly}
Zhekun Luo, Devin Guillory, Baifeng Shi, Wei Ke, Fang Wan, Trevor Darrell, and
  Huijuan Xu.
\newblock {Weakly-Supervised Action Localization with Expectation-Maximization
  Multi-Instance Learning}.
\newblock In {\em ECCV}, 2020.

\bibitem{ma2020sf}
Fan Ma, Linchao Zhu, Yi Yang, Shengxin Zha, Gourab Kundu, Matt Feiszli, and
  Zheng Shou.
\newblock {SF-Net: Single-Frame Supervision for Temporal Action Localization}.
\newblock In {\em ECCV}, 2020.

\bibitem{min2020adversarial}
Kyle Min and Jason~J Corso.
\newblock {Adversarial Background-Aware Loss for Weakly-Supervised Temporal
  Activity Localization}.
\newblock In {\em ECCV}, 2020.

\bibitem{narayan2021d2}
Sanath Narayan, Hisham Cholakkal, Munawar Hayat, Fahad~Shahbaz Khan, Ming-Hsuan
  Yang, and Ling Shao.
\newblock {D2-Net: Weakly-Supervised Action Localization via Discriminative
  Embeddings and Denoised Activations}.
\newblock In {\em ICCV}, 2021.

\bibitem{narayan20193c}
Sanath Narayan, Hisham Cholakkal, Fahad~Shahbaz Khan, and Ling Shao.
\newblock {3C-Net: Category Count and Center Loss for Weakly-Supervised Action
  Localization}.
\newblock In {\em ICCV}, 2019.

\bibitem{nguyen2018weakly}
Phuc Nguyen, Ting Liu, Gautam Prasad, and Bohyung Han.
\newblock {Weakly Supervised Action Localization by Sparse Temporal Pooling
  Network}.
\newblock In {\em CVPR}, 2018.

\bibitem{nguyen2019weakly}
Phuc~Xuan Nguyen, Deva Ramanan, and Charless~C Fowlkes.
\newblock {Weakly-Supervised Action Localization With Background Modeling}.
\newblock In {\em ICCV}, 2019.

\bibitem{pardo2021refineloc}
Alejandro Pardo, Humam Alwassel, Fabian Caba, Ali Thabet, and Bernard Ghanem.
\newblock {RefineLoc: Iterative Refinement for Weakly-Supervised Action
  Localization}.
\newblock In {\em WACV}, 2021.

\bibitem{paszke2019pytorch}
Adam Paszke, Sam Gross, Francisco Massa, Adam Lerer, James Bradbury, Gregory
  Chanan, Trevor Killeen, Zeming Lin, Natalia Gimelshein, Luca Antiga, et~al.
\newblock {PyTorch: An Imperative Style, High-Performance Deep Learning
  Library}.
\newblock {\em NeurIPS}, 2019.

\bibitem{paul2018w}
Sujoy Paul, Sourya Roy, and Amit~K Roy-Chowdhury.
\newblock {W-TALC: Weakly-supervised Temporal Activity Localization and
  Classification}.
\newblock In {\em ECCV}, 2018.

\bibitem{qing2021temporal}
Zhiwu Qing, Haisheng Su, Weihao Gan, Dongliang Wang, Wei Wu, Xiang Wang, Yu
  Qiao, Junjie Yan, Changxin Gao, and Nong Sang.
\newblock {Temporal Context Aggregation Network for Temporal Action Proposal
  Refinement}.
\newblock In {\em CVPR}, 2021.

\bibitem{ramezani2016review}
Mohsen Ramezani and Farzin Yaghmaee.
\newblock {A review on human action analysis in videos for retrieval
  applications}.
\newblock {\em Artificial Intelligence Review}, 2016.

\bibitem{shi2020weakly}
Baifeng Shi, Qi Dai, Yadong Mu, and Jingdong Wang.
\newblock {Weakly-Supervised Action Localization by Generative Attention
  Modeling}.
\newblock In {\em CVPR}, 2020.

\bibitem{shi2022dynamic}
Haichao Shi, Xiao-Yu Zhang, Changsheng Li, Lixing Gong, Yong Li, and Yongjun
  Bao.
\newblock {Dynamic Graph Modeling for Weakly-Supervised Temporal Action
  Localization}.
\newblock In {\em ACM MM}, 2022.

\bibitem{shou2016temporal}
Zheng Shou, Dongang Wang, and Shih-Fu Chang.
\newblock {Temporal Action Localization in Untrimmed Videos via Multi-Stage
  CNNs}.
\newblock In {\em CVPR}, 2016.

\bibitem{wang2021learning}
Deng-Bao Wang, Lei Feng, and Min-Ling Zhang.
\newblock {Learning from Complementary Labels via Partial-Output Consistency
  Regularization}.
\newblock In {\em IJCAI}, 2021.

\bibitem{wang2017untrimmednets}
Limin Wang, Yuanjun Xiong, Dahua Lin, and Luc Van~Gool.
\newblock {UntrimmedNets for Weakly Supervised Action Recognition and
  Detection}.
\newblock In {\em CVPR}, 2017.

\bibitem{xia2020part}
Xiaobo Xia, Tongliang Liu, Bo Han, Nannan Wang, Mingming Gong, Haifeng Liu,
  Gang Niu, Dacheng Tao, and Masashi Sugiyama.
\newblock {Part-dependent Label Noise: Towards Instance-dependent Label Noise}.
\newblock {\em NeurIPS}, 2020.

\bibitem{xu2020g}
Mengmeng Xu, Chen Zhao, David~S Rojas, Ali Thabet, and Bernard Ghanem.
\newblock {G-TAD: Sub-Graph Localization for Temporal Action Detection}.
\newblock In {\em CVPR}, 2020.

\bibitem{xu2019segregated}
Yunlu Xu, Chengwei Zhang, Zhanzhan Cheng, Jianwen Xie, Yi Niu, Shiliang Pu, and
  Fei Wu.
\newblock {Segregated Temporal Assembly Recurrent Networks for Weakly
  Supervised Multiple Action Detection}.
\newblock In {\em AAAI}, 2019.

\bibitem{yang2020revisiting}
Le Yang, Houwen Peng, Dingwen Zhang, Jianlong Fu, and Junwei Han.
\newblock {Revisiting Anchor Mechanisms for Temporal Action Localization}.
\newblock {\em TIP}, 29, 2020.

\bibitem{yang2021uncertainty}
Wenfei Yang, Tianzhu Zhang, Xiaoyuan Yu, Tian Qi, Yongdong Zhang, and Feng Wu.
\newblock {Uncertainty Guided Collaborative Training for Weakly Supervised
  Temporal Action Detection}.
\newblock In {\em CVPR}, 2021.

\bibitem{yang2022acgnet}
Zichen Yang, Jie Qin, and Di Huang.
\newblock {ACGNet: Action Complement Graph Network for Weakly-Supervised
  Temporal Action Localization}.
\newblock In {\em AAAI}, 2022.

\bibitem{yu2018learning}
Xiyu Yu, Tongliang Liu, Mingming Gong, and Dacheng Tao.
\newblock {Learning with Biased Complementary Labels}.
\newblock In {\em ECCV}, pages 68--83, 2018.

\bibitem{yuan2017temporal}
Zehuan Yuan, Jonathan~C Stroud, Tong Lu, and Jia Deng.
\newblock {Temporal Action Localization by Structured Maximal Sums}.
\newblock In {\em CVPR}, 2017.

\bibitem{zeng2019breaking}
Runhao Zeng, Chuang Gan, Peihao Chen, Wenbing Huang, Qingyao Wu, and Mingkui
  Tan.
\newblock {Breaking Winner-Takes-All: Iterative-Winners-Out Networks for Weakly
  Supervised Temporal Action Localization}.
\newblock {\em TIP}, 2019.

\bibitem{zeng2019graph}
Runhao Zeng, Wenbing Huang, Mingkui Tan, Yu Rong, Peilin Zhao, Junzhou Huang,
  and Chuang Gan.
\newblock {Graph Convolutional Networks for Temporal Action Localization}.
\newblock In {\em ICCV}, 2019.

\bibitem{zhai2020two}
Yuanhao Zhai, Le Wang, Wei Tang, Qilin Zhang, Junsong Yuan, and Gang Hua.
\newblock {Two-Stream Consensus Network for Weakly-Supervised Temporal Action
  Localization}.
\newblock In {\em ECCV}, 2020.

\bibitem{zhang2021cola}
Can Zhang, Meng Cao, Dongming Yang, Jie Chen, and Yuexian Zou.
\newblock {CoLA: Weakly-Supervised Temporal Action Localization With Snippet
  Contrastive Learning}.
\newblock In {\em CVPR}, 2021.

\bibitem{zhang2020multi}
Xiao-Yu Zhang, Haichao Shi, Changsheng Li, and Peng Li.
\newblock {Multi-Instance Multi-Label Action Recognition and Localization Based
  on Spatio-Temporal Pre-Trimming for Untrimmed Videos}.
\newblock In {\em AAAI}, 2020.

\bibitem{zhang2019learning}
Xiao-Yu Zhang, Haichao Shi, Changsheng Li, Kai Zheng, Xiaobin Zhu, and Lixin
  Duan.
\newblock {Learning Transferable Self-Attentive Representations for Action
  Recognition in Untrimmed Videos with Weak Supervision}.
\newblock In {\em AAAI}, volume~33, 2019.

\bibitem{zhao2017temporal}
Yue Zhao, Yuanjun Xiong, Limin Wang, Zhirong Wu, Xiaoou Tang, and Dahua Lin.
\newblock {Temporal Action Detection With Structured Segment Networks}.
\newblock In {\em ICCV}, 2017.

\bibitem{zhong2018step}
Jia-Xing Zhong, Nannan Li, Weijie Kong, Tao Zhang, Thomas~H Li, and Ge Li.
\newblock {Step-by-step Erasion, One-by-one Collection: A Weakly Supervised
  Temporal Action Detector}.
\newblock In {\em ACM MM}, 2018.

\bibitem{zhou2004multi}
Zhi-Hua Zhou.
\newblock {Multi-instance learning: A survey}.
\newblock {\em Department of Computer Science \& Technology, Nanjing
  University, Tech. Rep}, 2004.

\end{thebibliography}
}

\end{document}